\documentclass{llncs}
\usepackage[utf8]{inputenc}
\usepackage[table,xcdraw]{xcolor}
\usepackage{adjustbox}
\usepackage{nicefrac}
\usepackage{natbib}

\newcommand{\lb}{\cellcolor[HTML]{eeeeee}}
\newcommand{\mb}{\cellcolor[HTML]{bbbbbb}}

\usepackage{amssymb}

\usepackage{amsthm}
\usepackage{hhline}
\usepackage{algorithm2e}

\def\tu#1{\langle #1\rangle}
\newcommand{\A}{{\bf A}}
\newcommand{\I}{{\bf I}}
\newcommand{\B}{{\bf B}}
\newcommand{\E}{{\bf E}}

\newcommand{\X}{{$\times$}}

\title{From-Below Boolean Matrix Factorization Algorithm Based on MDL}

\author{%
	Tatiana Makhalova\inst{1,2}  \and 
	Martin Trnecka\inst{3} 
}

\institute{	
	National Research University Higher School of Economics, Moscow, Russia,
	LORIA, (CNRS -- Inria -- University of Lorraine), Vandœuvre-l\`es-Nancy, France\\
	Dept. Computer Science, Palacký University Olomouc, Olomouc, Czech Republic\\
	\email{tpmakhalova@hse.ru}, \email{martin.trnecka@gmail.com}
}

\begin{document}

\maketitle

\begin{abstract}
During the past few years Boolean matrix factorization (BMF) has become an important direction in data analysis. The minimum description length principle (MDL) was successfully adapted in BMF for the model order selection. Nevertheless, an BMF algorithm performing good results from the standpoint of standard measures in BMF is missing. In this paper, we propose a novel from-below Boolean matrix factorization algorithm based on formal concept analysis. The algorithm utilizes the MDL principle as a criterion for the factor selection. On various experiments we show that the proposed algorithm outperforms---from different standpoints---existing state-of-the-art BMF algorithms.

\end{abstract}


\section{Introduction}
\label{sec:introduction}

Boolean matrix factorization (BMF), also known as Boolean matrix decomposition, is a powerful and widely used data mining tool. Like a classical matrix factorization methods, e.g. non-negative matrix factorization (NNMF) or singular value decomposition (SVD), BMF provides a different description (see Section~\ref{sec:BMF}) of Boolean data, via new, more fundamental variables called factors.

In BMF a given input data matrix is approximated by a product of so-called object-factor and factor-attribute matrices. All matrices contain zeros and ones only. The quality of the factorization---i.e. the quality of factors themselves---is usually measured by standard measures in BMF, namely by the number of factors and by the coverage (how large is the portion of data is described by factors, see Section~\ref{sec:background}). Both can be easily implemented---in fact each susscefull BMF algorithm already utilized them---in an arbitrary BMF algorithm. Moreover, both are very important in the evaluation of the factorization quality~\cite{quality}. On the other hand, other aspects of the quality of factors, e.g. the interpretability, that are often neglected in the factor evaluation, are also an important parts of the matrix factorization.

By now, various approaches to assessment of the quality of factors were developed~\cite{quality,panda+}. One of the most fundamental---but surprisingly not often used---is based on the well-known minimum description length principle (MDL). In terms of MDL, the best factorization is the factorization with the minimal description. Due to the MDL principle, such factorization is useful and easily interpretable. Neverthless, it was many times shown (see, e.g.~\cite{quality,ess}) that mixing MDL and BMF produces a poor results with respect to the BMF standard error measures (the number of factors and the coverage). More details will be provided in Section~\ref{sec:background}.

Recent results~\cite{mdl-first} in the field of formal concept analysis (FCA)---which is related to the BMF (see Section~\ref{sec:fca})---involving the minimum description length (MDL) motivate us to revise the use of MDL in BMF.

We propose a new heuristic BMF algorithm for from-below matrix factorization that outperforms existing state-of-the-art algorithms and produces very good result w.r.t. the standard BMF measures. The algorithm utilizes formal concept analysis and the MDL principle. Additionally, we present an extensive experimental evaluation of factors delivered by the proposed algorithm and its comparison with some already existing algorithms.

The rest of the paper is organized as follows. In the following
Section~\ref{sec:related-work} we provide a brief overview of the related work. Then, in Section~\ref{sec:background}, a notation used in the paper, a short introduction to BMF and MDL, and a background of the paper are presented. Section~\ref{sec:alg} describes a design of our algorithm. The algorithm is experimentally evaluated in Section~\ref{sec:experiments}. Section~\ref{sec:conclusion} draws a conclusion and future research directions.

\section{Related Work}
\label{sec:related-work}

In the last decade, many BMF methods were developed~\cite{panda+,ess,asso,grecond,panda,hyper}. It was shown~\cite{dimension} that applying existing non Boolean methods (e.g. NNMF, SVD) on Boolean data is inappropriate, especially from the interpretation standpoint. 

A good overview of BMF and related topics can be found e.g. in \cite{quality,ess,asso}. In general, BMF and BMF algorithms are addressed in various papers involving formal concept analysis~\cite{grecond,bmf-ignatov}, role mining~\cite{exact}, binary databases~\cite{tiling} or bipartite graphs~\cite{biclique}.

In many application of BMF, instead of a general Boolean factorization---which can be computed for instance by well-known \textsc{Asso} algorithm---only a certain class of factorization, so-called from-below matrix factorization~\cite{ess}, is considered (see Section~\ref{sec:background}).

In the recent years, the minimum description length principle~\cite{mdl-book} has been applied in BMF. It was used mostly to solve the model order selection problem~\cite{model-selection}---i.e. separation of global structure from noise---or as a factor selection criteria in BMF algorithms, e.g. in the state-of-the-art algorithm \textsc{PaNDa$^+$}~\cite{panda+} (an improvement and generalized version of \textsc{PaNDa} algorithm~\cite{panda}). As a special case of  application of MDL in BMF \textsc{Hyper}~\cite{hyper} algorithm can be considered, its objective is to minimize the description of factors instead of the minimization of the description length (for more details see~\cite{panda+}).

Another related work is~\cite{mdl-first}, where a set of formal concepts with MDL is considered for the classification task. 
Our algorithm can be used for simillar tasks. Instead of~\cite{mdl-first} our algorithm does not require computing the whole set of formal concepts, that makes it applicable in practice. Moreover we used a different approach to MDL measuring.

This paper is, to the best of the author's knowledge, the first to address the from-below decomposition based on the MDL. 

\section{Background and Basic Definitions}
\label{sec:background}

\subsection{Notation}

Through the paper we use a matrix terminology and in some convenient places a relational terminology. Matrices are
denoted by upper-case bold letters ($\I$). $\I_{ij}$ denotes the entry corresponding to the row $i$ and the column $j$ of $\I$. The set of all $m \times n $ Boolean (binary) matrices is denoted by $\{0,1\}^{m\times n}$. The number of 1s in Boolean matrix $\I$ is denoted by $\|\I\|$, i.e $\|\I\| = \sum_{i,j} \I_{ij}$.

We interpret input data $\I \in \{0,1\}^{m\times n}$ primarily as an object-attribute incidence matrix, i.e. a relation between the set of objects and the set of attributes. That is, the entry $\I_{ij}$ is either $1$ or $0$, indicating that the object $i$ does or does not have the attribute $j$.

If ${\bf A} \in \{0,1\}^{m\times n}$ and ${\bf B} \in \{0,1\}^{m\times n}$, we have the following element-wise matrix operations. The {\it Boolean sum} ${\bf A} \oplus {\bf B}$ which is the normal matrix sum where $1+1 = 1$. The {\it Boolean subtraction} ${\bf A} \ominus {\bf B}$ which is the normal matrix subtraction, where $0 - 1 = 0$.

\subsection{Boolean Matrix Factorization}
\label{sec:BMF}

A general aim in BMF is for a given Boolean matrix $\I\in\{0,1\}^{m\times n}$
to find matrices $\A\in\{0,1\}^{m\times k}$ and $\B\in\{0,1\}^{k\times n}$ for which 

\begin{equation}
\label{eqn:model}
\I \approx \A \circ \B
\end{equation}

\noindent where $\circ$ is Boolean matrix multiplication, i.e.
$(\A\circ \B)_{ij} = \max_{l=1}^k \min(\A_{il},\B_{lj})$,
and $\approx$ represents approximate equality assessed by $||\cdot||$. The corresponding metric $E$ is defined for matrices $\I\in\{0,1\}^{m\times n}$, $\A\in\{0,1\}^{m\times k}$ and $\B\in\{0,1\}^{k\times n}$ by 

\begin{equation}
\label{eqn:error}
E(\I, \A \circ \B) = || \I \ominus (\A \circ \B) ||.
\end{equation}

A decomposition of $\I$ into $\A \circ \B$  may be interpreted as a discovery of $k$ factors that exactly or approximately explain the data: interpreting $\I$, $\A$, and $\B$ as the object–attribute, object–factor, and factor–attribute matrices, the model (\ref{eqn:model}) has the following interpretation: the object $i$ has the attribute $j$, i.e. $\I_{ij}=1$, if and only if there exists factor $l$ such that $l$ applies to $i$ and $j$ is one of the particular manifestations of $l$. 

Note also an important geometric view of BMF: a decomposition
$\I \approx \A \circ \B$  with $k$ factors represents a coverage
of the 1s in $\I$ by $k$ rectangular areas in $\I$ full of 1s, the $l$th rectangle is the Boolean sum of the $l$th
column in $\A$ and the $l$th row in $\B$. For more details see, e.g.~\cite{kim}.

If the rectangular areas cover only non zero elements in the matrix $\I$, the $\A \circ \B$ is called the {\it from-below matrix decomposition}~\cite{ess}. An example of the from-below BMF follows.

\begin{figure*}[ht]
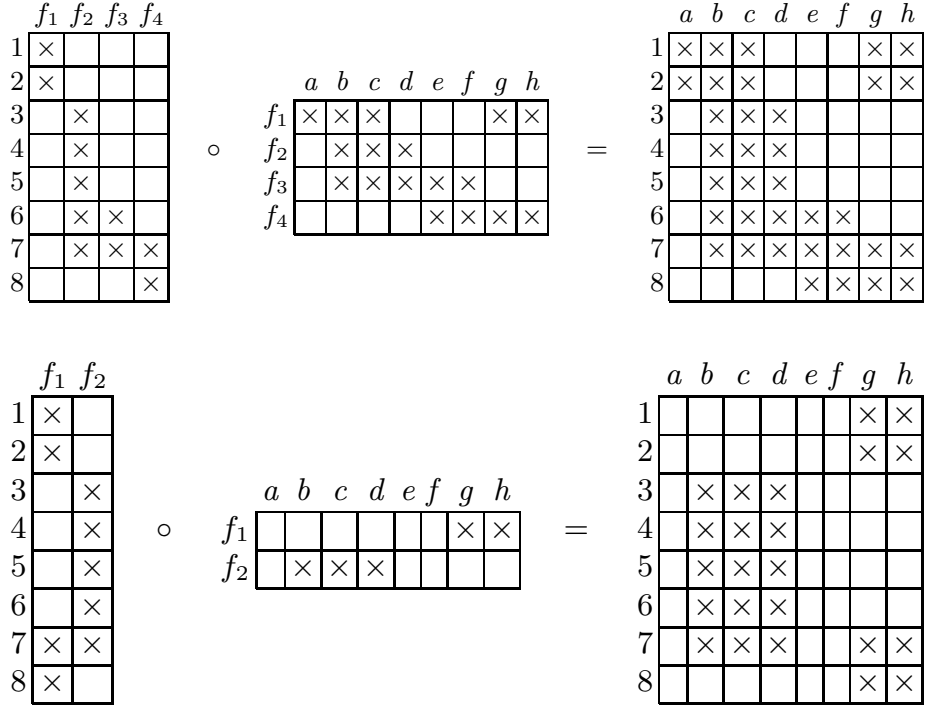

\small
\centering
\begin{adjustbox}{width=1\textwidth}
	\begin{tabular}{c|c|c|c|c|}
\multicolumn{1}{c}{}		& \multicolumn{1}{c}{$f_1$} & \multicolumn{1}{c}{$f_2$} & \multicolumn{1}{c}{$f_3$} & \multicolumn{1}{c}{$f_4$} \\ \cline{2-5}
		{1} & \X &  &  &  \\ \cline{2-5}
		{2} & \X &  &  &  \\ \cline{2-5}
		{3} &  & \X &  &  \\ \cline{2-5}
		{4} &  & \X &  &  \\ \cline{2-5}
		{5} &  & \X &  &  \\ \cline{2-5}
		{6} &  & \X & \X &  \\ \cline{2-5}
		{7} &  & \X & \X & \X \\ \cline{2-5}
		{8} &  &  &  & \X \\ \cline{2-5}
	\end{tabular}
 $\quad\circ\quad$ 
\begin{tabular}{c|c|c|c|c|c|c|c|c|}

	\multicolumn{1}{c}{}		& \multicolumn{1}{c}{\textit{a}} & \multicolumn{1}{c}{\textit{b}} & \multicolumn{1}{c}{\textit{c}} & \multicolumn{1}{c}{\textit{d}} & \multicolumn{1}{c}{\textit{e}} & \multicolumn{1}{c}{\textit{f}} & \multicolumn{1}{c}{\textit{g}} & \multicolumn{1}{c}{\textit{h}} \\ \cline{2-9}
		{$f_1$} & \X & \X & \X &  &  &  & \X & \X \\ \cline{2-9}
		{$f_2$} &  & \X & \X & \X &  &  &  &  \\ \cline{2-9}
		{$f_3$} &  & \X & \X & \X & \X & \X &  &  \\ \cline{2-9}
		{$f_4$} &  &  &  &  & \X & \X & \X & \X \\ 	\cline{2-9}
\end{tabular}
 $\quad=\quad$ 
\begin{tabular}{c|c|c|c|c|c|c|c|c|}
	\multicolumn{1}{c}{}		& \multicolumn{1}{c}{\textit{a}} & \multicolumn{1}{c}{\textit{b}} & \multicolumn{1}{c}{\textit{c}} & \multicolumn{1}{c}{\textit{d}} & \multicolumn{1}{c}{\textit{e}} & \multicolumn{1}{c}{\textit{f}} & \multicolumn{1}{c}{\textit{g}} & \multicolumn{1}{c}{\textit{h}} \\ \cline{2-9}
				{1} & \X & \X & \X &     &    &    & \X & \X\\ \cline{2-9}
				{2} & \X & \X & \X &    &    &    & \X & \X\\ \cline{2-9}
				{3} &    & \X & \X & \X &    &    &    &\\ \cline{2-9}
				{4} &    & \X & \X & \X &    &    &    &\\ \cline{2-9}
				{5} &    & \X & \X & \X &    &    &    & \\ \cline{2-9}
				{6} &    & \X & \X & \X & \X & \X &    &\\ \cline{2-9}
				{7} &    & \X & \X & \X & \X & \X & \X & \X\\ \cline{2-9}
				{8} &    &   &    &    & \X & \X & \X & \X\\ \cline{2-9}
			\end{tabular}
\end{adjustbox}
\bigskip
\vspace{0.8em}
\begin{adjustbox}{width=1.\textwidth}
\begin{tabular}{c|c|c|}
			\multicolumn{1}{c}{}	& \multicolumn{1}{c}{$f_1$} & \multicolumn{1}{c}{$f_2$} \\ \cline{2-3}
				{1} & \X &  \\ \cline{2-3}
				{2} & \X &  \\ \cline{2-3}
				{3} &  & \X \\ \cline{2-3}
				{4} &  & \X \\ \cline{2-3}
				{5} &  & \X \\ \cline{2-3}
				{6} &  & \X \\ \cline{2-3}
				{7} & \X & \X \\ \cline{2-3}
				{8} & \X &  \\ \cline{2-3}
			\end{tabular}
$\quad\circ\quad$
	\begin{tabular}{c|c|c|c|c|c|c|c|c|}
	\multicolumn{1}{c}{}		& \multicolumn{1}{c}{\textit{a}} & \multicolumn{1}{c}{\textit{b}} & \multicolumn{1}{c}{\textit{c}} & \multicolumn{1}{c}{\textit{d}} & \multicolumn{1}{c}{\textit{e}} & \multicolumn{1}{c}{\textit{f}} & \multicolumn{1}{c}{\textit{g}} & \multicolumn{1}{c}{\textit{h}} \\ \cline{2-9}
		{$f_1$} &  &  &  &  &  &  & \X & \X \\ \cline{2-9}
		{$f_2$} &  & \X & \X & \X &  &  &  &  \\ \cline{2-9}
	\end{tabular}
$\quad=\quad$
		\begin{tabular}{c|c|c|c|c|c|c|c|c|}
	\multicolumn{1}{c}{}		& \multicolumn{1}{c}{\textit{a}} & \multicolumn{1}{c}{\textit{b}} & \multicolumn{1}{c}{\textit{c}} & \multicolumn{1}{c}{\textit{d}} & \multicolumn{1}{c}{\textit{e}} & \multicolumn{1}{c}{\textit{f}} & \multicolumn{1}{c}{\textit{g}} & \multicolumn{1}{c}{\textit{h}} \\ \cline{2-9}
				{1} & & & &     & &   & \X & \X\\ \cline{2-9}
				{2} & & & &    &  &   & \X & \X\\ \cline{2-9}
				{3} & & \X & \X & \X &  &   &   &\\ \cline{2-9}
				{4} & & \X & \X & \X &  &  &   &\\ \cline{2-9}
				{5} & & \X & \X & \X &  &  &   & \\ \cline{2-9}
				{6} & & \X & \X & \X &  &  &   &\\ \cline{2-9}
				{7} & & \X & \X & \X &  &  & \X & \X\\ \cline{2-9}
				{8} & &   &    &    &  &  & \X & \X\\ \cline{2-9}
			\end{tabular}
\end{adjustbox}
\caption{Two examples of data factorization.}
\label{fig:example-data-factorization}
\end{figure*}

\begin{example}
\label{ex:BMF}
Let us consider Boolean matrix with rows $1, \dots, 8$ and columns $a, \dots, h$ depicted in Figure~\ref{fig:example-data}. The Boolean matrix is given in the shape of table, where nonzero entries are marked by crosses. Two different factorizations of the data are shown in Figure~\ref{fig:example-data-factorization}. 
\end{example}

\begin{figure}[ht!]
\small
\centering
\begin{adjustbox}{width=0.3\textwidth}
		\begin{tabular}{c|c|c|c|c|c|c|c|c|}	
	\multicolumn{1}{c}{}		& \multicolumn{1}{c}{\textit{a}} & \multicolumn{1}{c}{\textit{b}} & \multicolumn{1}{c}{\textit{c}} & \multicolumn{1}{c}{\textit{d}} & \multicolumn{1}{c}{\textit{e}} & \multicolumn{1}{c}{\textit{f}} & \multicolumn{1}{c}{\textit{g}} & \multicolumn{1}{c}{\textit{h}} \\ \cline{2-9}
			{1} & \X & \X & \X &  &  &  & \X & \X\\ \cline{2-9}
			{2} & \X & \X & \X &  &  &  & \X & \X\\ \cline{2-9}
			{3} & \X & \X & \X & \X &  &  &  &\\ \cline{2-9}
			{4} & \X & \X & \X & \X &  & \X &  &\\ \cline{2-9}
			{5} &  & \X & \X & \X &  &  &  & \X\\ \cline{2-9}
			{6} &  & \X & \X & \X & \X & \X &  &\\ \cline{2-9}
			{7} &  & \X & \X & \X & \X & \X & \X & \X\\ \cline{2-9}
			{8} &  &  &  & & \X & \X & \X & \X\\ \cline{2-9}
		\end{tabular}
\end{adjustbox}
\caption{Example data.}
\label{fig:example-data}
\end{figure}


\subsection{BMF with Help of Formal Concept Analysis}
\label{sec:fca}
{\it Formal concept analysis} (FCA)~\cite{fca} provides a basic framework for dealing with factors. The main notion of FCA is \emph{formal context}, which is usually represented as a Boolean matrix, it is defined as a triple $\left<\mathcal{X}, \mathcal{Y}, \mathcal{I} \right>$, where $\mathcal{X}$ is a nonempty set of objects , $\mathcal{Y}$ is a nonempty set of attributes and $\mathcal{I}$ is a binary relation between $\mathcal{X}$ and $\mathcal{Y}$. Hence the formal context $\left<\mathcal{X}, \mathcal{Y}, \mathcal{I} \right>$ with $m$ objects and $n$ attributes is a Booolean matrix  $\I\in\{0,1\}^{m\times n}$.

To every Boolean matrix $\I\in \{0,1\}^{n\times m}$, one might associate the pair 
$\tu{{}^{\uparrow}, {}^{\downarrow}}$ of
operators (in FCA well known as the arrow operators) assigning to sets 
$C\subseteq \mathcal{X}=\{1,\dots,m\}$ and $D\subseteq  \mathcal{Y}=\{1,\dots,n\}$ the sets $C^\uparrow\subseteq \mathcal{Y}$ and $D^\downarrow\subseteq \mathcal{X}$ defined by
\begin{eqnarray*}
  C^\uparrow = \{ j\in \mathcal{Y} \,|\, \forall i\in C: \I_{ij}=1\},\\
  D^\downarrow = \{ i\in \mathcal{X} \,|\, \forall j\in D: \I_{ij}=1\},
\end{eqnarray*}

where $C^\uparrow$ is the set of all attributes (columns) shared by all
objects (rows) in $C$ and $D^\downarrow$ is the set of all objects
sharing all attributes in $D$. 

The pair $\tu{C,D}$ for which $C^\uparrow = D$ and $D^\downarrow = C$ is called the \emph{formal concept}. $C$ and $D$  
are called the \emph{extent} and the \emph{intent} of formal concept $\tu{C,D}$, respectively. The concepts are partially ordered as follows: $\tu{A,B} \leq \tu{C,D}$ iff $A \subseteq C$ (or $D \subseteq B$), a pair $\tu{A,B}$ is a subconcept of $\tu{C,D}$, while $\tu{C,D}$ is a superconcept of $\tu{A,B}$. 
The set of all formal concepts we denote by
\[
    \mathcal{B}(\I) = \{ \tu{C,D} \mid C\subseteq \mathcal{X}, D\subseteq \mathcal{Y},
     C^\uparrow=D, D^\downarrow=C\}.
\]
The whole set of partially ordered formal concepts is called the \emph{concept lattice} of $\I$.

Given a set
${\cal F} = \{\tu{C_1,D_1},\dots,\tu{C_k,D_k}\} \subseteq {\cal B}(\I)$
(with a fixed indexing of the formal concepts $\tu{C_l,D_l}$),
induces the $m\times k$ and $k\times n$ 
Boolean matrices $\A_{\cal F}$ and $\B_{\cal F}$ by
\begin{eqnarray}\label{eqn:A}
   (\A_{\mathcal{F}})_{il}=\left\{
   \begin{array}{l}
     1, \textrm{if}~ i\in C_l,\\
     0, \textrm{if}~ i\not\in C_l,
   \end{array}
   \right.
\end{eqnarray}
and
\begin{eqnarray}\label{eqn:B}
   (\B_{\cal F})_{lj}=\left\{
   \begin{array}{l}
     1, \textrm{if}~ j\in D_l,\\
     0, \textrm{if}~ j\not\in D_l,
   \end{array}
   \right.
\end{eqnarray}
for $l=1,\dots,k$.
That is, the $l$th column and $l$th row of $\A_\mathcal{F}$ and $\B_\mathcal{F}$ are the characteristic vectors
of $C_l$ and $D_l$, respectively. The set $\mathcal{F}$ is also called a set of 
{\it factor concepts}. Clearly, $\A_\mathcal{F} \circ \B_\mathcal{F}$ is the from-below matrix decomposition.

\begin{example}
Let us considered two factorizations depicted in Figure~\ref{fig:example-data-factorization}. The first one corresponds to the set 
\begin{eqnarray*}
\mathcal{F} = \{\tu{\{1,2\}, \{a,b,c,g,h\}}, \tu{\{3,4,5,6,7\}, \{b,c,d\}},\\
\tu{\{6,7\}, \{b,c,d,e,f\}}, \tu{\{7,8\}, \{e,f,g,h\}}\}.
\end{eqnarray*}

\noindent The second one corresponds to the set 

\[\mathcal{F} = \{\tu{\{1,2,7,8\}, \{g,h\}}, \tu{\{3,4,5,6,7\}, \{b,c,d\}} \}.\]

\end{example}

For more details how formal concept analysis is utilized in BMF and the advantages of such approach see the pioneer work~\cite{grecond}.

\subsection{A Brief Introduction to MDL}

The minimum description length (MDL) principle, which is a computable
version of Kolmogorov complexity~\cite{mdl-book}, is a formalization of the law of parsimony, well known as Occam’s razor. In terms of MDL, it is formulated as follows: the best model is the model that ensures the best compression of the given data. 

More formally, for a given set of models $\mathcal{M}$ and data (in our case represented via Boolean matrix $\I$) the best model $M \in \mathcal{M}$ is the one that minimizes the following cost function: 

\begin{equation}
\label{eqn:MDL}
L(M) + L(\I \,|\, M),
\end{equation}

\noindent where $L(M)$ is the encoding length of  $M$ in bits and
$L(\I \,|\, M)$ is the encoding length  in bits of the data $\I$ encoded with $M$.

In general, we are only interested in the length
of the encoding, and not in the coding itself, i.e. we do not have
to materialize the codes themselves.

Note that MDL requires the compression to be lossless in order to allow for a
fair comparison between different models.

\subsection{The Quality of Factorization}
The quality of the obtained factorization~(\ref{eqn:model}) is usually evaluated via some variants of metric~(\ref{eqn:error}). From the BMF perspective there are two basic viewpoints, emphasizing the role of the first $k$ factors and the need to
account for a prescribed portion of data, respectively. They are
known as the discrete basis problem (DBP) and the approximate
factorization problem (AFP), see~\cite{asso} and \cite{grecond,ess}. Both of them emphasize the coverage of data, i.e. the geometric view of BMF. 

In many applications of BMF, the interpretation of factors plays a crucial role. It is reasonable instead of the coverage of the obtained factorization empathize a different quality measures that access the interpretability of factors, e.g. the MDL. 

On the other hand, the geometric view of BMF is very important and an interpretable factorization should reflect it.

In the next section, we propose a novel BMF algorithm which is based on well-known \textsc{GreConD} algorithm~\cite{grecond}. The algorithm computes from-below factorization via minimization of the cost function~(\ref{eqn:MDL}). The results of experiments show that it preserves a lot of information from the original data w.r.t. the error measure~(\ref{eqn:error}).

\section{Design of Algorithm}
\label{sec:alg}

\subsection{MDL in From-below Matrix Factorization}

For matrices $\A_\mathcal{F} \in \{0,1\}^{m\times k}$, $\B_\mathcal{F} \in \{0,1\}^{k\times m}$, and $\I \in \{0,1\}^{m\times n}$ where $\I \approx (\A_\mathcal{F} \circ \B_\mathcal{F})$ we define an error matrix $\E$ as follows:

\[
\I = (\A_\mathcal{F} \circ \B_\mathcal{F}) \oplus \E.
\]

One may observe that matrix $\E$ can be easily computed via metric~(\ref{eqn:error}), i.e. $\E = E(\I, \A_\mathcal{F} \circ \B_\mathcal{F})$. 
Hence, to provide a lossless compression of $\I$ it is sufficient to encode the matrices $\A_\mathcal{F}, \B_\mathcal{F}$ and $\E$, i.e. the MDL cost function~(\ref{eqn:MDL}) has the following form 

\begin{equation}
\label{eqn:mdl-bmf-model}
L(\A_\mathcal{F} \circ \B_\mathcal{F}) + L(\E).
\end{equation}

According to the MDL principle, the best factorization of $\I$ minimizes function (\ref{eqn:mdl-bmf-model}). In the following we explain how to compute the length of the encoding of matrices $\A_\mathcal{F}, \B_\mathcal{F}$ and $\E$ in bits. We use a similar approach as in~\cite{model-selection} and we modify it for the from-below matrix factorization.

More precisely, to use optimal prefix codes we need to encode the dimensions of the matrices and the matrices themselves, i.e.
\begin{eqnarray*}
L(\A_\mathcal{F} \circ \B_\mathcal{F}) + L(\E) &=& L(m) + L(n) + L(k) + \\
&  & +~ L(\A_\mathcal{F}) + L(\B_\mathcal{F}) + L(\E).
\end{eqnarray*}

\noindent For the sake of simplicity we may encode the dimensions $m, n, k$ with block-encoding, which give us $L(m) = L(n) = L(k) = \log(\max(m,n,k)).$

To not introduce some influencing between factors, these are encoded per factor, i.e. we encode $\A_\mathcal{F}$ per column and $\B_\mathcal{F}$ per row.

In order to use optimal prefix code, we need to first encode the probability of encountering 1 in a particular column or row respectively, i.e. we need $\log m$ bits for each extent in set $\mathcal{F}$ and $\log n$ bits for each $\mathcal{F}$ intent in set $\mathcal{F}$, respectively.

For simplicity, extent $C$ and intent $D$ of factor concept $\tu{C,D}$ can be seen as characteristic vectors, i.e. $C  \in \{0,1\}^{m\times 1}$ and $D  \in \{0,1\}^{1\times n}$. We need to encode all ones and zeros. The length of optimal code is determined by Shannon entropy. This gives us the number of bits required for the encoding of matrices $\A_\mathcal{F}$ and $\B_\mathcal{F}$:   
%

\begin{eqnarray*}
L(\A_\mathcal{F}) = \sum_{\tu{C,D} \in \mathcal{F}} \log m - (||C|| \cdot \log \frac{||C||}{m} + \\ 
+~(m-||C||) \cdot \log \frac{m - ||C||}{m}),
\end{eqnarray*}


%
%
%

\begin{eqnarray*}
L(\B_\mathcal{F}) = \sum_{\tu{C,D} \in \mathcal{F}} \log n - (||D|| \cdot  \log \frac{||D||}{n} + \\
+~(n-||D||) \cdot \log \frac{n - ||D||}{n}).
\end{eqnarray*}

In a similar way we can compute the number of bits required for the encoding of matrix $\E$:
%

\begin{eqnarray*}
L(\E) = \log mn - (||\E|| \cdot \log \frac{||\E||}{mn} +\\
+~ (mn-||\E||) \cdot \log \frac{mn - ||\E||}{ mn}).
\end{eqnarray*}

Note, we can encode matrix $\E$ element-by-element without any influence, because these elements are clearly independent.

\subsection{Algorithm}

In this section we propose a BMF algorithm, called \textsc{MDLGreConD}\footnote{\textsc{MDLGreConD} is an abbreviation of Minimum Description Length Greedy Concept on Demand.}, that uses the above described MDL cost function. 
The algorithm is a modified version---it utilizes a similar search strategy---of the \textsc{GreConD}\footnote{\textsc{GreConD} is an abbreviation of Greedy Concept on Demand.} algorithm~\cite{grecond}, which is one of the most successful from-below matrix decomposition algorithms (see e.g.~\cite{quality}). 

Pseudocode of \textsc{MDLGreConD} is depicted in Algorithm~\ref{alg:MDLGreConD}. The algorithm works as follows.

\begin{algorithm}[ht]
\DontPrintSemicolon
\LinesNumbered
\KwIn{Boolean matrix $\I$.}
\KwOut{Set $\mathcal{F}$ of factor concepts.}
\BlankLine

$\mathcal{F}\leftarrow \emptyset$\;
$total\_cost \leftarrow \infty$\;
$\E \leftarrow \I \ominus (\A_\mathcal{F} \circ \B_\mathcal{F})$\;

\While{$L(\A_\mathcal{F} \circ \B_\mathcal{F}) + L(\E)$ is decreasing}
{
	$\tu{C,D} \leftarrow \tu{\emptyset, \emptyset}$\;
	\While{$\tu{C,D}$ is changing}
	{
	$total\_cost' \leftarrow total\_cost$\;
	\ForEach{$j \notin D$}
	{
		$D' \leftarrow (D \cup \{j\})^{\downarrow\uparrow}$\;
		$C' \leftarrow D'^\downarrow$\;
		\If{$\tu{C',D'} \in \mathcal{F}$ }
		{
			continue with next $j$\;
		}
		
		$\mathcal{F'} \leftarrow \mathcal{F} \cup \tu{C', D'}$\;
		$cost \leftarrow L(\A_{\mathcal{F'}} \circ \B_{\mathcal{F'}}) + L(I \ominus (\A_{\mathcal{F'}} \circ \B_{\mathcal{F'}}))$
		
		\If{cost $<$ total\_cost'} 
		{
			$total\_cost' \leftarrow cost$\;
			$C'' \leftarrow C'$\;
			$D'' \leftarrow D'$\;
		}
			
	}
	$total\_cost \leftarrow total\_cost'$\;
				$C \leftarrow C''$\;
			$D \leftarrow D''$\;
	
	}
	
	$\mathcal{F} \leftarrow \mathcal{F} \cup \tu{C,D}$
}

\Return{$\mathcal{F}$}

\caption{\textsc{MDLGreConD} algorithm}
\label{alg:MDLGreConD}
\end{algorithm}

The algorithm computes a candidate $\tu{C,D}$ to a factor concept that minimizes the cost function~(\ref{eqn:mdl-bmf-model}) stored in variable {\it total\_cost}. This is done via searching of a promising column $j$ that is not included in $D$ (lines 8--21). Note that the adding of $j$ to $D$ is realized via ${}^\uparrow$ and ${}^{\downarrow}$ operators mentioned in Section~\ref{sec:fca}. Only the best column $j$ is considered (lines 16--20). If a new column is added to $\tu{C,D}$, i.e. the $\tu{C,D}$ is changed, the modified $\tu{C,D}$ is used as a new candidate and another promising column is searched for. If there is no column that reduce the cost function (line 6), already computed candidate is added to the output set $\mathcal{F}$ of factor concepts.
The algorithm ends if there is no candidate that allows for reduction of the cost function.

\subsection{Computational Complexity}
The Boolean matrix factorization problem is NP-hard~\cite{stockmeyer} as well as the computation of factorization that minimizes the cost function~(\ref{eqn:mdl-bmf-model}). The proposed algorithm is heuristic. One may easily derive an exact algorithm with an exponential time complexity. Such algorithm is inapplicable in practice. 

We do not provide the time complexity analysis, since the time complexity is not a main concern of Boolean matrix factorization. The presented algorithm is only slightly slower than \textsc{GreConD} algorithm, which is, probably, the fastest BMF algorithm  (see e.g.~\cite{ess}). 
Both of them are able to factorize, in order of second, on ordinary PC, all the data presented in Section~\ref{sec:experiments}.

\section{Experimental Evaluation}
\label{sec:experiments}

In this section, the results of an experimental comparison of BMF algorithms with \textsc{MDLGreConD} are presented.

\subsection{Datasets}

We use 6 different real-world datasets, namely {\tt Breast}, {\tt Ecoli}, {\tt Iris} and {\tt Mushroom} from UCI repository~\cite{uci}, and {\tt Domino} and {\tt Emea} from~\cite{exact}. The characteristics of the datasets are shown in Table~\ref{tab:datasets}. All of them are well known and widely used as benchmark datasets in BMF.

\begin{table}[ht]
\caption{Datasets and their characteristics.}
\smallskip 
\centering
\begin{tabular}{lr@{$\times$}lcr}
\hline
dataset		    & \multicolumn{2}{c}{size} & dens. $\I$ & $||\mathcal{B}(\I)||$ \\
\hline

{\tt Breast} 		& $699$	 	& $20$	& $0.499$  &  $642$\\
{\tt Domino} 		& $79$	 	& $231$	& $0.400$  &  $73$\\
{\tt Ecoli} 		& $336$	 	& $34$	& $0.235$  &  $813$\\
{\tt Emea} 		& $3046$	& $35$	& $0.068$  &  $780$\\
{\tt Iris} 		& $150$	 	& $19$	& $0.263$  &  $164$\\
{\tt Mushroom} 	& $8124$	& $90$	& $0.252$  &  $186332$\\

\hline
\end{tabular}
\label{tab:datasets}
\end{table}

\subsection{Algorithms}

\textsc{GreConD}~\cite{grecond} algorithm is based on the  ``on demand'' greedy
search  for formal concepts of $\I$.
It is designed to compute an exact from-below factorization. Instead of going through all formal concepts, which are the candidates for factor concepts, it constructs the factor concepts by adding sequentially ``promising columns'' to candidate $\tu{C,D}$ to factor concept. More formally, a new column $j$ that minimizes the error

\[E(\I,\A_{\mathcal{F} \cup \tu{(D \cup j)^{\downarrow},(D \cup j)^{\downarrow\uparrow}}}\circ \B_{\mathcal{F} \cup \tu{(D \cup j)^{\downarrow},(D \cup j)^{\downarrow\uparrow}}} )\] 

\noindent is added to $\tu{C,D}$. This is repeated until no such columns exist. If there is no such column, the $\tu{C,D}$ is added to the set $\mathcal{F}$. The algorithm ends if $E(\I,\A_\mathcal{F}\circ \B_\mathcal{F})$ is smaller than the prescribed parameter $\epsilon$ or the prescribed number of factors is reached. For more details see~\cite{ess}. Note, that usually $\epsilon = 0$, i.e. the whole matrix $\I$ is covered by factors. Such setting was adopted in our experiments.

\textsc{PaNDa$^+$} \cite{panda+} is an algorithmic framework based on \textsc{PaNDa}~\cite{panda} algorithm. The algorithm aims to extract a set $\mathcal{F}$ of pairs $\tu{C,D}$ that minimizes the cost function: 

\[\sum_{\tu{C,D}\in\mathcal{F}} (|C|+|D|) + E(\I,\A_\mathcal{F}\circ \B_\mathcal{F}).\] 

\noindent Every $\tu{C,D}$ in $\mathcal{F}$ is computed in two stages. On the first stage the core of  $\tu{C,D}$ is computed, on the second stage the core is extended. 
A core is a rectangle, not necessarily a formal concept, contained in $\I$ and it is computed by adding columns from a sorted list. 
Extension to $\tu{C,D}$ is performed by adding columns and rows to a core while such an addition allows for reducing  the cost. Note, that \textsc{PaNDa$^+$} does not produce the from-below factorization. The computation of \textsc{PaNDa$^+$} is driven by several parameters (see~\cite{panda+}). All of them are tuned for each dataset. The best obtained results are reported.

\textsc{Hyper}~\cite{hyper} algorithm aims to extract a set $\mathcal{F}$ of pairs $\tu{C,D}$ that minimize the cost function which is defined as follows: 

\[
\sum_{\tu{C,D}\in\mathcal{F}} (|C|+|D|) / E(\I,\A_\mathcal{F}\circ \B_\mathcal{F}).
\] 

\noindent As candidates to factors the set of all formal concepts $\mathcal{B}(\I)$ together with all single attribute rectangles in data are considered. Each candidate is divided into a set of single row rectangles that are sorted according to the number of uncovered elements in $\I$. Then the algorithm tries to add the single row rectangles back to the candidate, until the above mentioned cost function decreases. After this, the algorithm in each iteration selects the concept $\tu{C,D}$ from the modified set of candidates that minimizes the cost function. \textsc{Hyper} algorithm  produces the from-below factorization. The size of $\mathcal{B}(\I)$ can be exponentially large. In such case \textsc{Hyper} has the exponential time complexity. To reduce computational cost authors of \cite{hyper} propose to use only frequent formal concepts (the frequency is an additional parameter of the algorithm). Our experiments show that the frequency affects highly the performance of the algorithm. In our experiments we use the whole set of formal concepts $\mathcal{B}(\I)$, (for the set sizes see the last column of Table~\ref{tab:datasets}).

\subsection{Evaluation}

\label{sec:experiments-evaluation}

In our experiments we compare \textsc{MDLGreConD} algorithm with \textsc{GreConD}, \textsc{Hyper} and \textsc{PaNDa$^+$}. We study factors themselves and how well they cover the analyzed datasets.

\subsubsection{The number of factors} 
\label{sec:the-number-of-factors}
One of the main characteristic of BMF algorithms is the number of factors they produce. We measure not only the total number of factors, but also how many non-trivial factors are computed. Under trivial factors we mean the single-attribute ones. The results are shown in Table~\ref{tab:ntrivialfactors}. 

As it can be seen from the table, \textsc{PaNDa$^+$} tends to produce only few factors (w.r.t. the number of attributes, see Table~\ref{tab:datasets}).


\begin{table}[ht]
\caption{The number of factors.}
\centering
\begin{tabular}{llrr}

\hline
& & \multicolumn{2}{c}{no. of factors}\\
\cline{3-4}
dataset & algorithm & non-trivial & trivial\\
\hline

{\tt Breast} 
&\textsc{GreConD} & 15 & 4\\
&\textsc{Panda$^+$} &  4 &	0\\
&\textsc{Hyper} & 36 & 0 \\
&\textsc{MDLGreConD} & 6 & 1 \\
\hline

{\tt Domino} 
&\textsc{GreConD} & 13 & 8\\
&\textsc{Panda$^+$} &  3 & 0	\\
&\textsc{Hyper} & 10 &	132 \\
&\textsc{MDLGreConD} & 7 & 3 \\
\hline

{\tt Ecoli} 
&\textsc{GreConD} &  38 & 3 \\
&\textsc{Panda$^+$} & 6 & 0 \\
&\textsc{Hyper} & 35 & 30 \\
&\textsc{MDLGreConD} & 8 &	1 \\
\hline

{\tt Emea} 
&\textsc{GreConD} & 9 & 33 \\
&\textsc{Panda$^+$} & 3 & 0 \\
&\textsc{Hyper} & 3 & 35 \\
&\textsc{MDLGreConD} & 7 & 2 \\
\hline

{\tt Iris} 	
&\textsc{GreConD} & 8 & 12 \\
&\textsc{Panda$^+$} & 8 & 0 \\
&\textsc{Hyper} & 13 & 15 \\
&\textsc{MDLGreConD} & 7 & 0 \\
\hline

{\tt Mushroom}
&\textsc{GreConD} & 98 & 3 \\
&\textsc{Panda$^+$} & 8 & 0\\
&\textsc{Hyper} & 89 & 2 \\
&\textsc{MDLGreConD} & 50 & 0 \\

\hline
\end{tabular}
\label{tab:ntrivialfactors}
\end{table}

\textsc{Hyper} returns the number of factors which is close to the number of attributes. Moreover, more than a half of them are trivial. This is true on all datasets with an exception of {\tt Breast} and {\tt Mushroom} data. 

\begin{figure}[ht]
\centering
\includegraphics[width=0.75\linewidth]{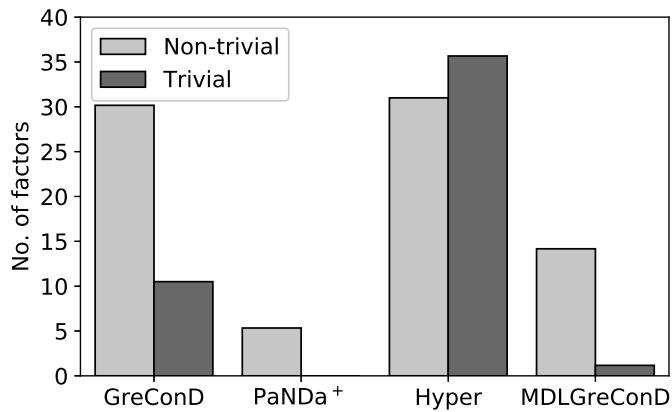}
\caption{The average number of factors.}
\label{fig:ntrivialfactors}
\end{figure}

On average (see Figure~\ref{fig:ntrivialfactors}), the number of non-trivial factors of \textsc{GreConD} is better than the number in case of \textsc{Hyper} algorithm. \textsc{MDLGreConD} generates a small set of factors, most of them are non-trivial. \textsc{PaNDa$^+$} tends to produce the smallest number of factors. All of them are non-trivial.

However, considering only the number of factors might be insufficient, since usually one wants to find not just the smallest number of factors, but the set of factors that capture (coverage) a large part of data. Further we will show how the algorithms capture the analyzed data.

\subsubsection{Data coverage} 
Another important characteristic of factors is how much information from the analyzed dataset they retain. We measure it by coverage rate. We differentiate \textit{data coverage} and \textit{object coverage}. Data coverage measures the rate of ``crosses'' covered by factors in the dataset---this is a standard measure in BMF, see e.g.~\cite{quality}. However, data coverage might be an inappropriate measure in cases where a dataset contains a lot of redundant attributes. Taking into consideration these cases, we measure the object coverage rate, i.e. how many objects are covered at least by one factor. The following example explains how the coverage measures are computed. 

\begin{example} 
The factor set of the first factorization (Figure~\ref{fig:example-data-factorization}) covers almost all crosses in data, while the second set covers around a half of crosses. The coverings for both of them are given below. The crosses covered by one factor are light gray, the crosses covered by more factors are colored with darker gray.

\begin{figure*}[ht]
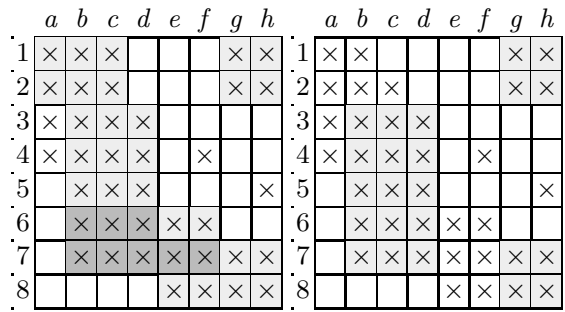

\centering
\begin{adjustbox}{width=.6\textwidth}
	
	\begin{tabular}{c|c|c|c|c|c|c|c|c|}
		\multicolumn{1}{c}{}		& \multicolumn{1}{c}{\textit{a}} & \multicolumn{1}{c}{\textit{b}} & \multicolumn{1}{c}{\textit{c}} & \multicolumn{1}{c}{\textit{d}} & \multicolumn{1}{c}{\textit{e}} & \multicolumn{1}{c}{\textit{f}} & \multicolumn{1}{c}{\textit{g}} & \multicolumn{1}{c}{\textit{h}} \\ 
	\hhline{*{1}{|~}*{8}{|-}|}
	{1} &\lb \X &\lb \X &\lb \X &  &  &  & \lb \X & \lb \X\\ \hhline{*{1}{|~}*{8}{|-}|}
	{2} &\lb \X &\lb \X &\lb \X &  &  &  & \lb \X & \lb \X\\ \hhline{*{1}{|~}*{8}{|-}|}
	{3} & \X &\lb \X & \lb \X &\lb \X &  &  &  & \\ \hhline{*{1}{|~}*{8}{|-}|}
	{4} & \X &\lb \X & \lb \X &\lb \X &  & \X &  & \\ \hhline{*{1}{|~}*{8}{|-}|}
	{5} &  &\lb \X &\lb \X & \lb \X &  &  &  & \X\\ \hhline{*{1}{|~}*{8}{|-}|}
	{6} &  &\mb \X & \mb \X &\mb \X & \lb \X & \lb \X &  &\\ \hhline{*{1}{|~}*{8}{|-}|}
	{7} &  &\mb \X & \mb \X &\mb \X & \mb \X &\mb \X & \lb \X &\lb \X\\ \hhline{*{1}{|~}*{8}{|-}|}
	{8} &  &  &  &  &\lb \X & \lb \X &\lb \X & \lb \X\\ \hhline{*{1}{|~}*{8}{|-}|}
	\end{tabular}

	\begin{tabular}{c|c|c|c|c|c|c|c|c|}
		\multicolumn{1}{c}{}		& \multicolumn{1}{c}{\textit{a}} & \multicolumn{1}{c}{\textit{b}} & \multicolumn{1}{c}{\textit{c}} & \multicolumn{1}{c}{\textit{d}} & \multicolumn{1}{c}{\textit{e}} & \multicolumn{1}{c}{\textit{f}} & \multicolumn{1}{c}{\textit{g}} & \multicolumn{1}{c}{\textit{h}} \\ 
	\hhline{*{1}{|~}*{8}{|-}|}
	{1}&\X&\X&&&&&\lb \X&\lb \X\\ \hhline{*{1}{|~}*{8}{|-}|}
	{2}&\X&\X&\X&&&&\lb \X&\lb \X\\ \hhline{*{1}{|~}*{8}{|-}|}
	{3}&\X&\lb \X&\lb \X&\lb \X&&&& \\ \hhline{*{1}{|~}*{8}{|-}|}
	{4}&\X&\lb \X&\lb \X&\lb \X&&\X&& \\ \hhline{*{1}{|~}*{8}{|-}|}
	{5}&&\lb \X&\lb \X&\lb \X&&&&\X \\ \hhline{*{1}{|~}*{8}{|-}|}
	{6}&&\lb \X&\lb \X&\lb \X&\X&\X&&\\ \hhline{*{1}{|~}*{8}{|-}|}
	{7}&&\lb \X&\lb \X&\lb \X&\X&\X&\lb \X&\lb \X\\ \hhline{*{1}{|~}*{8}{|-}|}
	{8}&&&&&\X&\X&\lb \X&\lb \X\\ \hhline{*{1}{|~}*{8}{|-}|}
	\end{tabular}
\end{adjustbox}
\caption{The covering with factors from the running examples.}
\label{fig:example_overlapping}
\end{figure*}
	
Note, both factor sets cover all objects, i.e. every row in the dataset has at least one colored cross, thus the object coverage rates is equal to 1 for both factorizations.

For the first factorization, the cross coverage rate is $\nicefrac{35}{39} = 0.897$. In the case of the second factorization, the cross coverage rate is $\nicefrac{23}{39} = 0.589$. Obviously, the bigger value is better.

\end{example}

\begin{figure}[ht]
	\centering
	\includegraphics[width=0.65\linewidth]{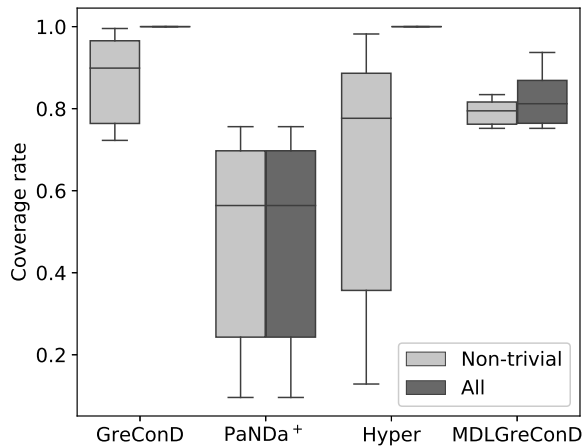}
	\caption{The average data coverage rate.}
	\label{fig:coveragerate-crosses}	
\end{figure}

\begin{figure}[ht]
	\centering
	\includegraphics[width=0.65\linewidth]{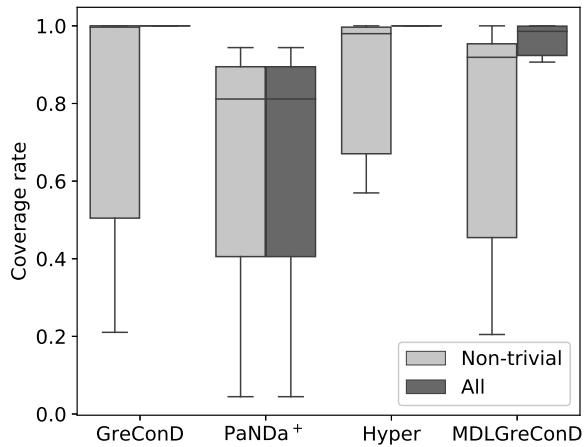}
	\caption{The average object coverage rate.}
	\label{fig:coveragerate-object}	
\end{figure}

Average values of data coverage and object coverage rates over all datasets as well as the minimal, maximal values and quantiles are shown in Figures~\ref{fig:coveragerate-crosses} and \ref{fig:coveragerate-object} respectively. The average data coverage rate of non-trivial factors of \textsc{MDLGreConD} is slightly lower than the analogous measure for \textsc{GreConD} and \textsc{Hyper}. It is important to note that \textsc{MDLGreConD} provides more stable results, in other words, the data coverage rate does not depend a lot on datasets, while for \textsc{Hyper} algorithm, the data coverage rate changes from 0.2 to 1.0. \textsc{PaNDa$^+$} covers slightly more than a half of data by a small set of factors. Moreover, if we take into account results regarding the number of factors from Section~\ref{sec:the-number-of-factors} \textsc{MDLGreConD} outperforms all remaining algorithms. Namely, it provides a large coverage by a smaller number of factors.

Regarding the object coverage rate, all the algorithms have similar performance, however a large number of non-trivial factors in \textsc{Hyper} ensures its high coverage rate for all chosen datasets.

\subsubsection{Redundancy of factors} 
An important characteristic of a factor set is redundancy. The factor set is redundant if it contains repetitive information, i.e. if it contains some overlaps between factors. We measure redundancy by {\it overlapping rate} (see Example~\ref{ex:overlap}), i.e. how many times the covered crosses are covered by several factors. 

\begin{example}
\label{ex:overlap}
For the factor sets from Figure~\ref{fig:example-data-factorization} the average overlapping rate is computed as follows.  We count the total area of factors $area(\tu{C,D}) = ||C|| \cdot ||D||$. In the case of the first factorization we obtain $area(f_1) = 10$, $area(f_2) = 15$, $area(f_3) = 10$ and $area(f_4) = 8$. The total area is 43, the number of covered crosses is 35, thus, the average overlapping rate is $\nicefrac{43}{35}$. The second factorization is without overlapped crosses, thus its average overlapping rate is 1.
\end{example}

\begin{figure}
	\centering
	\includegraphics[width=0.65\linewidth]{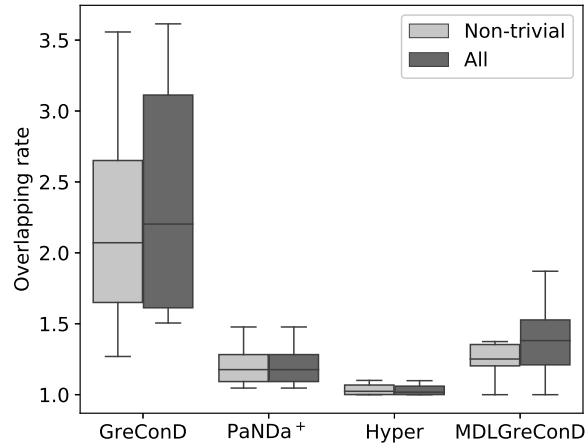}
	\caption{The average overlapping rate.}
	\label{fig:overlapping-rate}
\end{figure}

Averages values of overlapping rate are shown in Figure~\ref{fig:overlapping-rate}. Our experiments show that factor sets with minimal redundancy are produced by \textsc{Hyper} algorithm. It can be explained regarding the previous experiments (see Section~\ref{sec:the-number-of-factors}), where it was shown that \textsc{Hyper} algorithm tends to produce a large number of trivial factors. \textsc{PaNDa$^+$} tends to produce a very small number of factors with low coverage rate. As one may clearly observe, \textsc{GreConD} produces factorizations with the largest overlapping rate. \textsc{MDLGreConD} generates a non-redundant set.

\subsection{Discussion}
Let us summarize the experimental evaluation. \textsc{GreConD} and \textsc{Hyper} are both able to explain the whole data. However, the quality of factorizations they produce is lower than the quality of \textsc{MDLGreConD}. More precisely, \textsc{Hyper} produces a large number of trivial factors. \textsc{GreConD} produce a less number of trivial factors, but with a lot of overlappings between them.

The quality of factorization obtained via \textsc{PaNDa$^+$} algorithm is low as well. The factors delivered by \textsc{PaNDa$^+$} cover only a small part of input data.

According to the experimental evaluation, \textsc{MDLGreConD} algorithm provide a factor set with well-balanced characteristics. The number of factors is reasonably small, factors themselves explain a large portion of data and are not redundant.

\section{Conclusions}
\label{sec:conclusion}
In this paper an MDL-based from-below factorization algorithm, which utilizes formal concept analysis, has been proposed. It produces a small subset of formal concepts having a low information loss rate.

The proposed algorithm does not require computing the whole set of formal concepts, that makes it applicable in practice. More than that, it computes factor sets that have better overall characteristics than factor sets computed by the existing BMF algorithms. The \textsc{MDLGreConD}-generated factor sets are small, contain few single-attribute factors and have a high coverage with low overlapping rate. 

An important direction of future work is application of the proposed method under supervised settings, i.e. for dealing with classification tasks.

\bibliographystyle{plain}

\end{document}